\pdfoutput=1
\documentclass[11pt]{article}

\usepackage{ACL2023} % [review]

\usepackage{times}
\usepackage{latexsym}
\usepackage[T1]{fontenc}
\usepackage[utf8]{inputenc}

\usepackage{microtype}

\usepackage{inconsolata}
\usepackage{graphicx} 
\usepackage{multirow}
\usepackage{tabularx}
\usepackage {amssymb}
\usepackage{arydshln}

\title{PetKaz at SemEval-2024 Task 8: \\ Can Linguistics Capture the Specifics of LLM-generated Text?}

\author{Kseniia Petukhova,  Roman Kazakov, Ekaterina Kochmar\\
  Mohamed bin Zayed University of Artificial Intelligence \\
  \texttt{\{kseniia.petukhova, roman.kazakov, ekaterina.kochmar\}@mbzuai.ac.ae} \\}

\begin{document}
\maketitle
\begin{abstract}
In this paper, we present our submission to the SemEval-2024 Task~8 ``Multigenerator, Multidomain, and Multilingual Black-Box Machine-Generated Text Detection'', focusing on the detection of machine-generated texts (MGTs) in English. Specifically, our approach relies on combining embeddings from the RoBERTa-base with diversity features and uses a resampled training set. We score 12th from 124 in the ranking for Subtask A (monolingual track), and our results show that our approach is generalizable across unseen models and domains, achieving an accuracy of 0.91. Our code is available at \url{https://github.com/sachertort/petkaz-semeval-m4}.
% In this work, we present our project focusing on the detection of machine-generated texts (MGTs), which we do within the framework of participation in SemEval-2024 Task~8, titled ``Multigenerator, Multidomain, and Multilingual Black-Box Machine-Generated Text Detection''. 
% We propose a comprehensive feature set that includes stylometric features, lexical diversity measures, descriptive statistics, readability scores, rhetorical structure theory (RST) parsing, and entity grid discourse representations. We conducted several experiments to test different configurations of these features. We focused on enhancing a pre-trained model, which was originally provided as a baseline by the task organizers. By incorporating our extracted features into this model via a feed-forward neural network, we were able to outperform the original baseline model. Code is available at \url{https://github.com/sachertort/petkaz-semeval-m4/tree/main}. Our name on the leaderboard %\footnote{\url{https://www.codabench.org/competitions/1752/#/results-tab}}
% of the task is \texttt{sachertort}.
\end{abstract}

\section{Introduction}

SemEval-2024 Task~8 ``Multigenerator, Multidomain, and Multilingual Black-Box Machine-Generated Text Detection''~\citep{wang-EtAl:2024:SemEval20245} has focused on the detection of machine-generated texts (MGTs). In recent years, large language models (LLMs) have achieved human-level performance across multiple tasks, showing impressive capabilities in natural language understanding and generation~\cite{llms-a-survey}, including their abilities to generate high-quality content in such areas as news, social media, question-answering forums, educational, and even academic contexts. Often, text generated by LLMs is almost indistinguishable from that written by humans, especially along such dimensions as text fluency~\citep{mitchell2023detectgpt}. Therefore, methods of automated MGT detection, intending to mitigate potential misuse of LLMs, are quickly gaining popularity. Automated MGT detection methods can be roughly split into black-box and white-box types, with the former being restricted to API-level access to LLMs and reliant on features extracted from machine-generated and human-written text samples for classification model training, and the latter focusing on zero-shot AI text detection without any additional training (see Section \ref{sec:rel}).

\begin{figure}
  \centering
  \includegraphics[width=\linewidth]{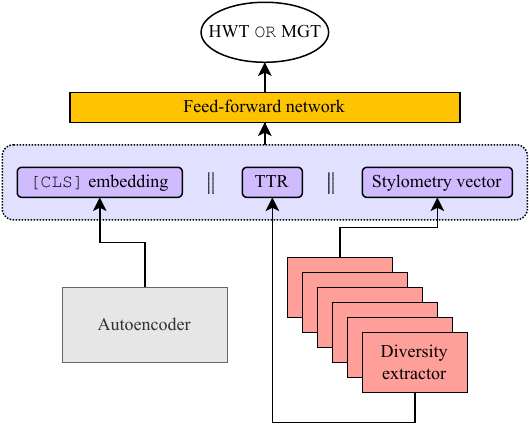}
  \vspace{-0.9cm}
  \caption{For each text, we get a \texttt{[CLS]} token embedding from an autoencoder model and extract vectors of linguistic features (e.g., lexical diversity, stylometry, etc.). Then, we pass the concatenated vector to a feed-forward network, whose output layer performs binary classification -- HWT vs. MGT. The configurations of embeddings/features may vary between experiments.}
  \label{fig:scheme}
  % \vspace{-0.2cm}
\end{figure}

For our submission to SemEval-2024 Task~8, the monolingual track of Subtask A, which focuses on MGT detection in English across a variety of domains and generative models, we have developed a system that can be categorized as a black-box detector and is based on a combination of embeddings, measures of lexical diversity, and careful selection of the training data (see Figure~\ref{fig:scheme}). We also present and discuss an extended set of linguistic features, including discourse and stylistic features, that we have experimented with during the development phase of the competition. The main motivation for using such a feature-based approach is that it helps us to focus on the fundamental differences between MGTs and human-written texts (HWTs) rather than capture the specifics of particular models.

Our results suggest that our best model, which uses diversity features and embeddings, outperforms a very competitive baseline introduced in this task~\cite{wang-EtAl:2024:SemEval20245}, yielding an accuracy of 0.95 on the development and 0.91 on the test set. It brought us 12th place out of 124 teams participating in the shared task. Furthermore, our investigation shows that a model using no embeddings but relying on such linguistic features as entity grid and stylometry yields results that are on par with the baseline model.

The main contributions of our work are as follows: {\bf (1)} we investigate the impact on the detection task of a variety of linguistically motivated features, ranging from widely used stylometric features to novel ones, including those based on high-level discourse analysis; and {\bf (2)} we show how training data can be selected in an informative way to help models better distinguish between MGTs and HWTs.

\section{Related Work}
\label{sec:rel}

A comprehensive survey by \citet{yang2023survey} categorizes detection methods into training-based classifiers, zero-shot detectors, and watermarking techniques, covering both black-box and white-box detection scenarios. This survey discusses a range of strategies, including mixed training, proxy models, and semantic embeddings, indicating ongoing challenges in scalability and robustness. Given the fast development of LLMs and their capabilities, of particular interest are innovations in zero-shot detection methods highlighted by~\citet{mitchell2023detectgpt} and~\citet{su2023detectllm}. In addition,~\citet{mitchell2023detectgpt} present DetectGPT, utilizing perturbation discrepancies to discern MGTs, while \citet{su2023detectllm} propose DetectLLM-LRR and DetectLLM-NPR, which advance zero-shot detection by harnessing log rank information.

Another relevant line of research investigates the use of linguistic and stylometric features, such as the ones overviewed in~\citet{bergsma2012stylometric}, for MGT detection. For instance, \citet{wang2023m4} explore the use of logistic regression with GLTR features (analyzing the distribution of token probabilities and their relative frequencies within specific probability ranges from a language model's output), stylistic characteristics, and NELA news verification features (style, complexity, bias, affect, morality, and event specifics) on the M4 dataset, and~\citet{liu2022coco} introduce a model exploiting text coherence, named entities and relation-aware graph convolutional networks under a low-resource setting for MGT detection.

\section{Methodology}
Our general pipeline, visualized in  Figure~\ref{fig:scheme}, consists of the following components: (1) an autoencoder model fine-tuned on HWT vs. MGT classification task; (2) linguistic features extraction pipeline; and (3) embeddings and features combination passed through a feed-forward neural network. Below we describe some of these components in more detail.

\subsection{Embeddings}
% As a baseline, we use \texttt{roberta-base}\footnote{\url{https://huggingface.co/FacebookAI/roberta-base}} model. 
We employ an autoencoder model. First, we fine-tune it on the HWT vs. MGT classification task, and then we use its \texttt{[CLS]} tokens' embeddings in a feed-forward model.

\subsection{Features}
\label{sec:feat}
We study the impact on the classification accuracy of several types of linguistically motivated features extracted from texts, including those based on: 1)~text statistics; 2)~readability; 3)~stylometry; 4)~lexical diversity; 5)~rhetorical structure theory (RST); and 6)~entity grid. Below we provide a description of the features and their relevance to the task. 

\paragraph{Text statistics}
We compute the following:\footnote{We use Python's \texttt{textstat} library: \url{https://pypi.org/project/textstat/}.} 1)~the number of difficult words (words that have more than two syllables and are not in the list of easy words\footnote{\url{https://github.com/textstat/textstat/blob/main/textstat/resources/en/easy_words.txt}} from~\citealp{e5c6a2f3-a448-300c-922c-4e2ddd10f19a}); 2)~raw lexicon count (unique words in text); 3)~raw sentence count. In Appendix~\ref{sec:textstats}, we provide the values for HWTs and across models.

\paragraph{Readability}
We assess the readability of MGTs and HWTs guided by the hypothesis that HWTs are easier to read than MGTs. We calculate a range of common readability scores for both types of texts to assess their readability, including 1)~Flesch Reading Ease Test~\citep{Flesch1979HowTW}; 2)~Flesch-Kincaid Grade Level Test~\citep{Kincaid1975DerivationON}; and 3)~Linsear Write Metric~\citep{o1966gobbledygook}.

\paragraph{Stylometry}
For stylometric features, we use the approach proposed in~\citet{bergsma2012stylometric}. Specifically, we collect all unigrams and bigrams from the texts and keep punctuation, stopwords, and Latin abbreviations (e.g., \textit{i.e.}) unchanged. Then, we build two types of representations where other words are replaced by their PoS tags and ``spelling signatures'' (forms of words; e.g., \texttt{xXX-dd} for \textit{iOS-17}).\footnote{The pre-processing was done using \texttt{spaCy}: \url{https://spacy.io}.}
Then, log token frequencies (TFs) are computed for each text and passed to the maximum absolute scaler, and these sparse representations are used as features. For further processing, sparse matrices with stylometry features are reduced by truncated singular value decomposition to a dimensionality of 768. See Appendix~\ref{sec:styleimp} for the analysis of stylometry features importance.

\paragraph{Lexical diversity}
Lexical diversity tells us how ``rich'' texts are in terms of vocabulary, i.e., whether they use rare words, or include a wide range of synonyms, epithets, terms, etc. There are a few measures widely used to measure lexical diversity, mostly based on the variants of the type-token ratio (TTR). We extract 10 features, such as TTR, Maas TTR, Hypergeometric distribution $d$~\citep[HDD;][]{mccarthy2007vocd}, etc.\footnote{Using Python's \texttt{lexical\_diversity} library: \url{https://github.com/kristopherkyle/lexical_diversity}.} For an in-depth overview, see~\citeposs{mccarthy2010mtld} study on lexical diversity assessment.

\paragraph{RST features}
In rhetorical structure theory (RST), proposed in~\citet{mann1988rhetorical}, texts are analyzed in terms of hierarchical structures, which represent the organization of information and text flow. These structures are made up of elementary discourse units (EDUs) connected through rhetorical relations, which include ``elaboration'', ``contrast'', ``cause'', ``result'', etc. Using an open-source sentence-level RST parser~\citep{lin2019unified}, we count the occurrences of various relations in each text and divide them by the total number of sentences in the text. 

\paragraph{Entity grid}
Finally, we use the entity grid algorithm to analyze the coherence of text by capturing patterns of entity distribution~\citep{barzilay2008modeling}. This method transforms a text into sequences of entity transitions, documenting the distribution, syntax, and reference information of discourse entities. Entities from texts are first tagged with their syntactic roles\footnote{Noun coreference is resolved using \texttt{spaCy} (\url{https://spacy.io}) and \texttt{neuralcoref} (\url{https://spacy.io/universe/project/neuralcoref}).} and categorized into three types: subject (\texttt{s}), object (\texttt{o}), and other (\texttt{x}).
The next step involves examining the transition of entities' roles across consecutive sentence pairs. This includes transitions like subject-to-object, object-to-other, subject-to-none, among others. Finally, we calculate the frequency of each transition type for all entities by dividing the total count of each transition type by the number of sentence pairs.

\subsection{Feed-forward neural network}
Finally, we use a concatenation of embeddings and vectors representing combinations of various features described above and pass them as input to a feed-forward neural network. Then, the output layer performs binary classification.

\section{Data}

Shared task organizers have used an extension of the M4 dataset~\citep{wang2023m4},\footnote{\url{https://github.com/mbzuai-nlp/M4}} which covers a range of domains (including {\em WikiHow}, {\em Wikipedia}, {\em Reddit}, {\em arXiv}, {\em PeerRead}, and {\em Outfox}) and texts generated by a number of LLMs (including {\tt ChatGPT}, {\tt Cohere}, {\tt Davinci003}, {\tt Dolly-v2}, {\tt BLOOMZ}, and {\tt GPT-4}) as well as written by humans. 
%In Appendix~\ref{sec:datastats}, we present detailed statistics on the data distribution across models and domains in each subset.
Overall, the training set is roughly balanced between HWTs and MGTs, with 53\% being HWTs and with the number of HWTs being around 5 times higher than that of texts generated by any single LLM for each of the domains. The only exception is {\em PeerRead}, where the distribution of texts generated by each LLM and written by humans is about the same. At the same time, the distribution is exactly 50\%:50\% for HWTs:MGTs in the development set, and 47.5\%:52.5\% for HWTs:MGTs in the test set. In addition, while both training and development sets cover a range of domains, the test set is limited to {\em Outfox} only. % (see Appendix \ref{sec:datastats}). 

\paragraph{A curious case of WikiHow}
\label{sec:wikihow}

Before running the experiments, we further investigate how the training data is composed. According to~\citet{wang2023m4}, LLMs were provided with relatively short inputs to generate texts across various domains: for example, with titles for {\em Wikipedia} articles and {\em arXiv} papers, titles and abstracts for {\em PeerRead} articles, etc. On the one hand, we observe a high level of parallelism in the training data across HWTs and texts generated by various models, and on the other, we note that there is little consistency in what models generate in certain domains: for example, provided with a name of a personality they generate quite different  \emph{Wikipedia} entries, which do not only differ from the correspondent HWTs but also vary from one LLM to another (see examples in Appendix~\ref{sec:datastats}, Table~\ref{tab:parallel_train}). In contrast, texts in the {\em WikiHow} domain appear to be more similar to each other across LLMs, which can be explained either by the way the data was generated (using titles and headlines as prompts to produce MGTs) or by the fact that there are fewer ways to explain {\em How to do X?} compared to the tasks in other domains. Moreover, our experiments with in-domain training of the MGT detection classifier suggest that the best results can be obtained when it is trained on the {\em WikiHow} domain. 
We follow up on these observations and create a customized training subset by using all MGTs from the original data and limiting HWTs to the texts from the {\em WikiHow} domain only. This results in a training set of 56,406 MGTs and 15,499 HWTs, with the distribution between each LLM and humans being roughly 1:1.

%It can be assumed that the output of an LLM writing a paper abtsract or a Wikipedia article on the basis of just the paper / article titles would be considerably different from the original human-written abstract or article. This prevents any substantial parallelism between HWTs and MGTs. On the other hand, provided a question on {\em How to do X?} (for WikiHow, they chose randomly 3,000 articles with lengths of more than 1,000 characters and used their titles and headlines as prompts to produce new machine-generated articles) an LLM might generate text that is consistent with and structurally and semantically very similar to that written in response to the same question by a human (after all, how many way are there to buy a guinea pig cage?) This hypothesis of parallelism present in WikiHow but not elsewhere needs to be tested further. 

%It looks like the data in the test set is also parallel to a large extent, while that in the development set shows less parallelism. We conclude that training classifiers on such parallel data is beneficial, and perform data selection limiting the training data to that coming from {\em WikiHow}.

\section{Experiments}

\subsection{Experimental setup}
As the source of embeddings, we use \texttt{roberta-base}\footnote{\url{https://huggingface.co/FacebookAI/roberta-base}}~\citep{liu2019roberta} fine-tuned within the baseline framework\footnote{\url{https://github.com/mbzuai-nlp/SemEval2024-task8/tree/main/subtaskA/baseline}} over 3 epochs with the learning rate of 2$e$-5 and $L_2$ norm of the weights being 0.01. 
The feed-forward neural network with two hidden layers accompanied by a ReLU activation function is then trained with the learning rate 5$e$-5, $L_2$ norm of the weights 0.01, and early stopping after 25 epochs. Each hidden layer has batch normalization and a dropout of 0.5.
We use \texttt{PyTorch}\footnote{\url{https://pytorch.org}}~\citep{paszke2019pytorch} for all training and evaluation steps. 

Following up on our observations on the {\em WikiHow} subset described in Section~\ref{sec:wikihow}, we conduct two series of experiments and train the feed-forward network on: 1) the {\em full} training set; and 2) the {\em reduced} training set where we use MGTs from all domains and HWTs from {\em WikiHow} only.

\subsection{Experiments on the development set}

\begin{table}
    \centering
    \begin{tabular}{lcc}\hline
         \textbf{Configuration} & \textbf{Full train} & \textbf{Reduced train}\\ \hline
         \texttt{feat} & 0.60 & 0.60\\
         \texttt{sty} & 0.68 & 0.57\\
         \texttt{sty} $\mathbin\Vert$ \texttt{feat} & 0.69 & 0.60\\
         \texttt{sty} $\mathbin\Vert$ \texttt{div} & 0.65 & 0.72\\
         \texttt{sty} $\mathbin\Vert$ \texttt{read} & 0.67 & 0.61\\
         \texttt{sty} $\mathbin\Vert$ \texttt{rst} & 0.64 & 0.57\\
         \texttt{sty} $\mathbin\Vert$ \texttt{ent} & 0.73 & 0.56\\
         \texttt{emb} & 0.74 & 0.83\\
         \texttt{emb} $\mathbin\Vert$ \texttt{sty} & 0.73 & 0.82\\
         \texttt{emb} $\mathbin\Vert$ \texttt{feat} & 0.76 & 0.90\\
         \texttt{emb} $\mathbin\Vert$ \texttt{div} & 0.73 & \textbf{0.95}\\
         \texttt{emb} $\mathbin\Vert$ \texttt{read} & 0.72 & 0.81\\
         \texttt{emb} $\mathbin\Vert$ \texttt{rst} & 0.73 & 0.81\\
         \texttt{emb} $\mathbin\Vert$ \texttt{ent} & 0.73 & 0.82\\ \hdashline
         Baseline & 0.74 & --\\
         \hline
    \end{tabular}
    \caption{Accuracy of different configurations and the baseline on the development set. \texttt{feat} stands for all features except stylometry, \texttt{sty} -- stylometry, \texttt{div} -- lexical diversity, \texttt{read} -- text statistics and readability, \texttt{rst} -- RST, \texttt{ent} -- entity grid, \texttt{emb} -- embeddings (see Section \ref{sec:feat}).}
    \label{tab:metrics}
\end{table}

The evaluation results of our model with different feature configurations applied to the development set are presented in Table~\ref{tab:metrics}. Several observations are due at this point.

First of all, we note that the highest accuracy of 0.95 is achieved with the model trained on the {\em reduced} training set using a combination of embeddings and diversity features. This does not mean that lexical diversity is necessarily the most powerful among linguistic features, but it suggests that it complements embedding representations better than other linguistic features. Moreover, it is the only feature type that increases the accuracy obtained with embeddings only. Finally, we also note that with the linguistic features, our model can outperform a competitive baseline used by the task organizers, which sets the accuracy at 0.74.

Secondly, stylometry features turn out to be the best linguistic feature type when used on their own: the accuracy with \texttt{sty} is 0.68 vs. 0.6 with \texttt{feat}. These representations reflect some general patterns of word types used in texts. However, it seems like they alone are not enough for effective classification, at least when applied to texts generated by modern LLMs. 
Notably, the configuration that combines stylometry with entity grid features (\texttt{sty + ent}) demonstrates performance that is nearly identical to the baseline employing a pre-trained language model (0.73 vs. 0.74), suggesting that entity grid adds further information about text coherence. Other features like RST do not seem to help distinguish MGTs from HWTs. This finding suggests that the frequency or efficacy with which humans and models employ rhetorical structures is comparable.

Finally, we observe that the performance of the model using {\tt emb} features always increases if it is trained on the {\em reduced} set. This determines the model configuration for our final submission.

\section{Results}
\begin{table}
    \centering
    \begin{tabular}{lccc} \hline
        \textbf{Configuration} & \textbf{Train} & \textbf{Accuracy} & \textbf{F\textsubscript{1}} \\ \hline
         \texttt{emb} $\mathbin\Vert$ \texttt{div} & reduced & \textbf{0.91} & \textbf{0.92} \\
         \texttt{sty} $\mathbin\Vert$ \texttt{ent} & full & 0.84 & 0.85\\ \hdashline
         Baseline & full & 0.88 & -- \\ \hline
    \end{tabular}
    \caption{Metrics on the test set. The first row is our main submitted configuration. The organizers do not report only the baseline's $F_1$ score.}
    \label{tab:test}
\end{table}

Table~\ref{tab:test} presents accuracy on the test set obtained with two configurations: a model using embeddings and lexical diversity features trained on the reduced training set, and a model using stylometry and entity grid features trained on the full training set, which showed promising results on the development set. \textbf{The former one is our main configuration: our team has submitted its predictions for the test set and scored 12th in the shared task (out of 124 teams).} This model outperforms the organizers' baseline, which sets the  accuracy at 0.88. However, we note that the latter model, which relies on linguistic features only and does not employ any pre-trained language model, also shows promising results, further strengthening our hypothesis that linguistic features are able to capture important properties of LLM-generated texts.

\subsection{Analysis}
\begin{figure}[t]
  \vspace{-0.25cm}
  \centering
  \includegraphics[width=\linewidth]{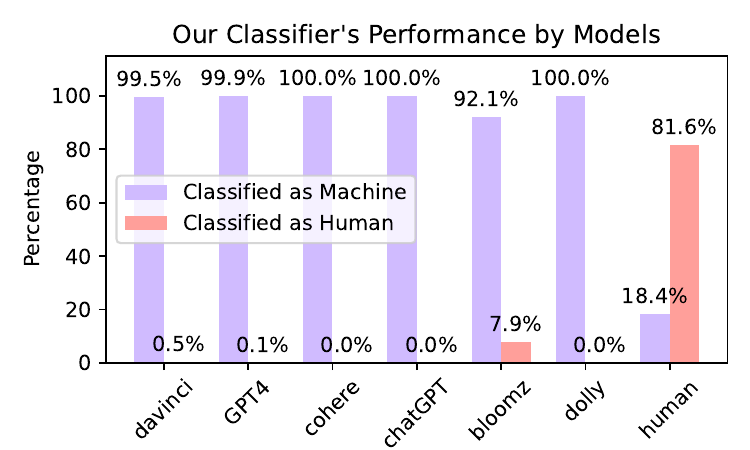}
  \vspace{-0.8cm}
  \caption{Performance of our classifier across models.}
  \label{acrossmodels}
\end{figure}

We further analyze the performance of our best model across different LLMs on the test set, as illustrated in Figure~\ref{acrossmodels}. The results show that our model accurately identifies texts from \texttt{Dolly-v2}, \texttt{Cohere}, and \texttt{ChatGPT} as machine-generated, and achieves near-perfect classification precision on texts from \texttt{GPT-4} and \texttt{Davinci003}. \texttt{BLOOMZ} is the only model that presents a problem for our classifier, with an 8\% misclassification rate. Additionally, we observe that 18\% of HWTs are incorrectly classified as being generated by machines. This shows the remarkable generalizability of our approach compared to~\citet{wang2023m4}, who reported that ``it is challenging for detectors to generalize well on unseen examples if they are either from different domains or are generated by different large language models. In such cases, detectors tend to misclassify machine-generated text as human-written''.

\begin{figure}[t]
  \centering
  \includegraphics[width=\linewidth]{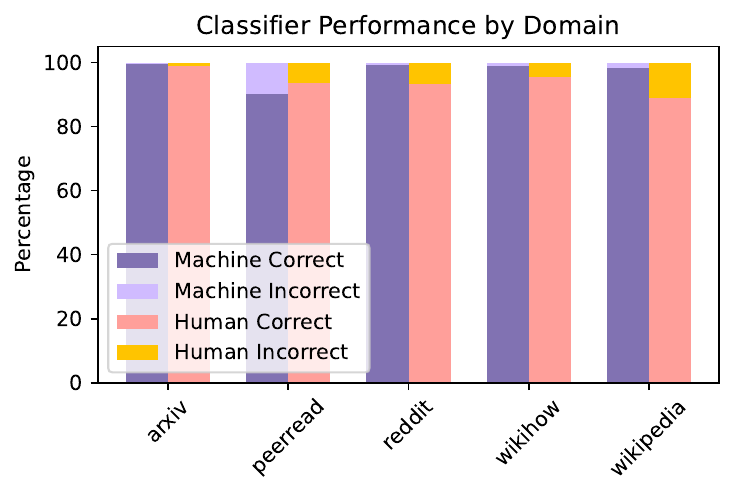}
  \vspace{-1cm}
  \caption{Performance of our classifier across domains (on the development set).}
  \label{acrossdomains}
\end{figure}

Furthermore, we evaluate our model's performance across domains (Figure~\ref{acrossdomains}). Our analysis reveals that we can accurately identify all MGTs and nearly perfectly recognize HWTs from \emph{arXiv}. Our classifiers face the biggest difficulties when classifying MGTs from \emph{PeerRead} and HWTs from \emph{Wikipedia}. These results are aligned with those reported in~\citet{wang2023m4}, who also found that training on \emph{Wikipedia} leads to the worst out-of-domain accuracy.

In summary, our classifier demonstrates generalizability, performing well on both previously unseen models (\texttt{GPT-4} and \texttt{BLOOMZ}) and domains (with all texts in the test set being from \emph{Outfox}).

\section{Conclusions}

When developing the models for our submission to the SemEval-2024 Task~8, we have primarily focused on: (1) the contribution of linguistic features to the task, and (2) the selection of the informative training data. Our results suggest that models using only linguistic features (specifically, those based on stylometry and entity grid) can perform competitively on this task, while careful selection of the training data helps improve the performance of the models that rely on embeddings. This shared task demonstrates that it is possible to distinguish between HWTs and MGTs, but the results also suggest promising avenues for future research, including in-depth analysis of the training data selection techniques and expansion of the linguistic features.

\section*{Limitations}

Our work is limited to the English language only as we opted to participate in a single Subtask of SemEval-2024 Task 8. In addition, this work is only limited to the domains and LLMs included in the shared task data, therefore, the generalizability of our approach beyond these domains and LLMs will need to be verified in future experiments.

\section*{Acknowledgements}
We would like to express our gratitude to Ted~Briscoe for inspiring us with the idea that linguistics could be of help and for engaging in discussions with us. 
We are grateful to Mohamed bin Zayed University of Artificial Intelligence (MBZUAI) for supporting this work. We also thank the anonymous reviewers for their valuable feedback.

\bibliography{anthology,custom}
\bibliographystyle{acl_natbib}

\newpage
\appendix

\section{Features Analysis}
\subsection{Text statistics across models}
\label{sec:textstats}
Table~\ref{tab:descr} shows various text statistics calculated on the training set. It can be seen that HWTs have higher values than all MGTs across all these metrics.

\begin{table}[ht]
    \centering
    \begin{tabular}{lccc} \hline 
         \textbf{Model}&  \textbf{DW}&  \textbf{LC}& \textbf{SC}\\ \hline 
         {\tt ChatGPT}&  64&  350& 19\\ 
         {\tt Cohere}&  37&  256& 13\\ 
         \texttt{Davinci003}&  58&  315 & 16\\ 
         {\tt Dolly-v2} &  54&  342 & 18\\  
         Human & 91& 582 & 30\\ \hline
    \end{tabular}
    \caption{Text statistics on the training set. DW~=~difficult words (mean), LC~=~lexicon count (mean), SC~=~sentence count (mean).}
    \label{tab:descr}
\end{table}

\subsection{Stylometry features importance}
\label{sec:styleimp}
Stylometry features are passed to a linear SVM classifier\footnote{From \texttt{scikit-learn}~\citep{scikit-learn}: \url{https://scikit-learn.org}.} to extract coefficients that may be interpreted as feature importances. Table~\ref{tab:sty} presents the most important features for MGTs and HWTs in the case of binary classification: for example, we can see that proper nouns are mostly associated with HWTs. It also makes it clear how the features are ordered by importance.

\begin{table}[ht]
    \centering
    \begin{tabular}{lc}
    \hline
         \textbf{MGT feature} & \textbf{Wt.} \\ \hline
         \textit{How to} & 3.28 \\
         \textsc{space} \textit{How}& 2.34 \\
         \textsc{num verb}& 2.07 \\
         \texttt{Xxxxx} \textit{the}& 2.00 \\
         \textit{How}& 1.78 \\
         \textsc{space num} & 1.77 \\
         \textit{Well}& 1.57 \\
         \texttt{Xxx} \textit{the}& 1.38 \\
         \texttt{dd Xxxxx}& 1.37 \\
         \textsc{noun} \textit{you}& 1.37 \\\hline
    \end{tabular}
    \begin{tabular}{lc}
    \hline
         \textbf{HWT feature} & \textbf{Wt.}\\ \hline
         \textsc{noun space} & -4.12\\
         \textsc{space} & -4.12\\
         \texttt{xxxx} & -3.93\\
         \textsc{space adj} & -3.10\\
         \textsc{space propn} & -3.10\\
         \textit{the} \textsc{space} & -2.67\\
         \textsc{noun} & -2.57\\
         \textsc{num space} & -2.21\\
         \textsc{propn space} & -2.15\\
         \texttt{\_XXX\_d} & -2.14\\\hline
    \end{tabular}
    \caption{Stylometric features highly weighted by the binary SVM classifier.}
    \label{tab:sty}
\end{table}

\section{Data Statistics}
\label{sec:datastats}

% \begin{table*}[ht]
%     \centering
%     \begin{tabular}{lrrrrrr} \hline 
%          \textbf{Model}& \textbf{\emph{WikiHow}}&  \textbf{\emph{Wikipedia}}& \textbf{\emph{Reddit}} & \textbf{\emph{arXiv}}& \textbf{\emph{PeerRead}} & {\bf Total}\\ \hline 
%          {\tt ChatGPT}& 3,000 & 2,995 & 3,000 & 3,000 & 2,344 & 14,339 \\ 
%          {\tt Cohere}& 3,000 & 2,336 & 3,000 & 3,000 & 2,342 & 13,678 \\ 
%          \texttt{Davinci003}& 3,000 & 3,000 & 3,000 & 2,999 &  2,344 & 14,343 \\ 
%          {\tt Dolly-v2} & 3,000 & 2,702 & 3,000 & 3,000 & 2,344 & 14,046 \\  
%          Human & 15,499 & 14,497 & 15,500 & 15,498 & 2,357 & 63,351 \\ \hline
%          {\bf Total} & 27,499 & 25,530 & 27,500 & 27,497 & 11,731 & 119,757 \\ \hline
%     \end{tabular}
%     \caption{Domain and model distribution in the training set.}
%     \label{tab:descr_train}
% \end{table*}

% \begin{table*}[ht]
%     \centering
%     \begin{tabular}{lrrrrrr} \hline 
%          \textbf{Model}&  \textbf{\emph{WikiHow}}&  \textbf{\emph{Wikipedia}}& \textbf{\emph{Reddit}} & \textbf{\emph{arXiv}}& \textbf{\emph{PeerRead}} & {\bf Total}\\ \hline 
%          {\tt BLOOMZ}& 500 & 500 & 500 & 500 & 500 & 2,500 \\ 
%          Human & 500 & 500 & 500 & 500 & 500 & 2,500 \\ \hline
%          {\bf Total} & 1,000 & 1,000 & 1,000 & 1,000 & 1,000 & 5,000 \\ \hline
%     \end{tabular}
%     \caption{Domain and model distribution in the development set.}
%     \label{tab:descr_dev}
% \end{table*}

\begin{table*}[ht]
    \centering
    \resizebox{\linewidth}{!}{
    \begin{tabular}{ll} \hline 
    \multicolumn{2}{c}{ {\bf \em WikiHow}} \\ \hline
    {\tt ChatGPT}& Buying Virtual Console games for your Nintendo Wii is a fun and easy process that can net \\ 
    & you some classic games to play on your console. [...]  \\
    {\tt Cohere}& How to Buy Virtual Console Games for Nintendo Wii \\ 
    & The Nintendo Wii has a feature called the Virtual Console that allows you to download and \\
    & play games from past Nintendo consoles, such as the Nintendo Entertainment System. [...]  \\
    {\tt Davinci003}& How to Buy Virtual Console Games for Nintendo Wii \\
    & Most people know that Nintendo's library of classic titles is available on the Wii platform \\
    & through the Virtual Console. [...]  \\
    {\tt Dolly-v2}& Find a few Wii Points cards from game retailers like GameStop., Make sure your Wii is  \\
    & online and on a secure connection if possible. [...]  \\
    Human& They are about \$20 a card. Or, if you want to just buy points with your credit card, Skip  \\
    & down to the section, With a Credit Card. [...]  \\
    \hline   

    \multicolumn{2}{c}{ {\bf \em Wikipedia}} \\ \hline
    {\tt ChatGPT}& William Whitehouse was a 19th-century British engineer and inventor who made significant \\
    & contributions to the field of hydraulics. [...]  \\
    {\tt Cohere}& William Whitehouse (1567-1648) was an English scholar, schoolmaster, and Anglican \\
    & clergyman. [...] \\
    {\tt Davinci003}& William Whitehouse (August 6, 1590 - May 18, 1676) was an English priest, scholar and \\
    & biblical commentator. [...]  \\
    {\tt Dolly-v2}& William Whitehouse (born William John Whitehouse; 15 July 1944) is an English musician, \\
    & singer and songwriter.  [...]  \\
    Human& William Edward Whitehouse (20 May 1859 - 12 January 1935) was an English cellist. [...]  \\
    \hline 

    \multicolumn{2}{c}{ {\bf \em PeerRead}} \\ \hline
    {\tt ChatGPT}& The paper 
    "End-to-End Learnable Histogram Filters" aims to introduce a novel approach \\
    & that enables histogram filters to be learnable end-to-end. [...]  \\
    {\tt Cohere}& This paper addresses the problem of designing end-to-end learnable histogram filters. [...]  \\
    {\tt Davinci003}& This paper presents an interesting approach to combining problem-specific algorithms with \\
    & machine learning techniques to find a balance between data efficiency and generality. [...]  \\
    {\tt Dolly-v2}& The paper End-to-End Learnable Histogram Filters demonstrates an interesting technique \\
    & for reducing photo noise without blurring the image. [...]  \\
    Human& We are retracting our paper "End-to-End Learnable Histogram Filters" from ICLR to submit \\
    & a revised version to another venue. [...]  \\
    \hline 
   
    \end{tabular}
    }
    \caption{Parallel HWTs and texts generated by different LLMs in the training set extracted from selected domains.}
    \label{tab:parallel_train}
\end{table*}

% Tables~\ref{tab:descr_train} and ~\ref{tab:descr_dev} show domain and model distribution in M4 training and development sets used in the shared task, respectively. Training set consists of 56,406 MGTs and 63,351 HWTs, showing class distribution close to but (not exactly equal to) 50\%:50\%. At the same time, the dvelopment set contains 2,500 MGTs from a single LLMs and 2,500 HWTs, keeping the proportion exactly 50\%:50\%. Finally, test set consists of 16,272 HWTs and 18,000 MGTs generated by various models (including some new ones, such as {\tt GPT-4}) and originating from a single domain ({\em Outfox}).

% In addition, 
Table~\ref{tab:parallel_train} shows some examples of parallel texts extracted from three domains represented in the training set ({\em WikiHow}, {\em Wikipedia}, and {\em PeerRead}). As explained in~\citet{wang2023m4}, the data for each domain was generated in a different way (for instance, using an article title only in some cases, and more extended inputs in others). We observe that there is much less consistency between the outputs generated by different LLMs in such domains as {\em Wikipedia} and {\em PeerRead} than in {\em WikiHow}. For instance, in the case of generated {\em Wikipedia} articles, the models cannot even agree on what personality they are describing (which is obvious from the very first sentences of such generated articles), while in the case of generated reviews from {\em PeerRead}, article descriptions also exhibit high diversity in the way they are presented in the review. At the same time, we hypothesize that generating texts for the {\em WikiHow} domain, describing {\em How to do X?}, results in higher consistency in the models' outputs, which is exemplified in Table~\ref{tab:parallel_train}.

%\begin{table*}[ht]
%    \centering
%    \begin{tabular}{|lrrrrr|r|} \hline 
%         \textbf{Model}&  \textbf{WikiHow}&  \textbf{Wikipedia}& \textbf{Reddit} & \textbf{ArXiv}& \textbf{PeerRead} & {\bf Total}\\ \hline 
%         {\tt ChatGPT}& 3,000 & 2,995 & 3,000 & 3,000 & 2,344 & 14,339 \\ 
%         {\tt Cohere}& 3,000 & 2,336 & 3,000 & 3,000 & 2,342 & 13,678 \\ 
%         \texttt{Davinci003}& 3,000 & 3,000 & 3,000 & 2,999 &  2,344 & 14,343 \\ 
%         {\tt Dolly-v2} & 3,000 & 2,702 & 3,000 & 3,000 & 2,344 & 14,046 \\  
%         Human & 15,499 & 14,497 & 15,500 & 15,498 & 2,357 & 63,351 \\ \hline
%         {\bf Total} & 27,499 & 25,530 & 27,500 & 27,497 & 11,731 & 119,757 \\ \hline
%    \end{tabular}
%    \caption{Domain and model distribution in the test set.}
%    \label{tab:descr_test}
%\end{table*}

\end{document}